\pdfoutput=1

\documentclass[11pt]{article}

\usepackage[preprint]{main}

\usepackage{times}
\usepackage{latexsym}

\usepackage[T1]{fontenc}

\usepackage[utf8]{inputenc}

\usepackage{microtype}

\usepackage{inconsolata}

\usepackage{graphicx}
\usepackage{array}
\usepackage{booktabs} 
\usepackage{algorithm}
\usepackage{algpseudocode}
\usepackage{multirow}
\usepackage{amssymb}
\usepackage{pifont}
\usepackage{makecell}
\usepackage{subfigure}
\title{Promoting Data and Model Privacy in Federated Learning \\ through Quantized LoRA}

\author{JianHao Zhu, Changze Lv, Xiaohua Wang, Muling Wu, Wenhao Liu, \\ {\bf Tianlong Li, Zixuan Ling, Cenyuan Zhang, Xiaoqing Zheng\thanks{\ \ Corresponding author.}, Xuanjing Huang} \\
School of Computer Science, Fudan University, Shanghai, China \\
\texttt{\{zhujh22\}@m.fudan.edu.cn} \\
\texttt{\{zhengxq,xjhuang\}@fudan.edu.cn} \\}

\begin{document}
\maketitle
\begin{abstract}

Conventional federated learning primarily aims to secure the privacy of data distributed across multiple edge devices, with the global model dispatched to edge devices for parameter updates during the learning process. 
However, the development of large language models (LLMs) requires substantial data and computational resources, rendering them valuable intellectual properties for their developers and owners.
To establish a mechanism that protects both data and model privacy in a federated learning context, we introduce a method that just needs to distribute a quantized version of the model's parameters during training. 
This method enables accurate gradient estimations for parameter updates while preventing clients from accessing a model whose performance is comparable to the centrally hosted one. 
Moreover, we combine this quantization strategy with LoRA, a popular and parameter-efficient fine-tuning method, to significantly reduce communication costs in federated learning. 
The proposed framework, named \textsc{FedLPP}, successfully ensures both data and model privacy in the federated learning context. 
Additionally, the learned central model exhibits good generalization and can be trained in a resource-efficient manner.

\end{abstract}
\begin{figure*}[htbp]
    \centering    \includegraphics[width=0.95\linewidth]{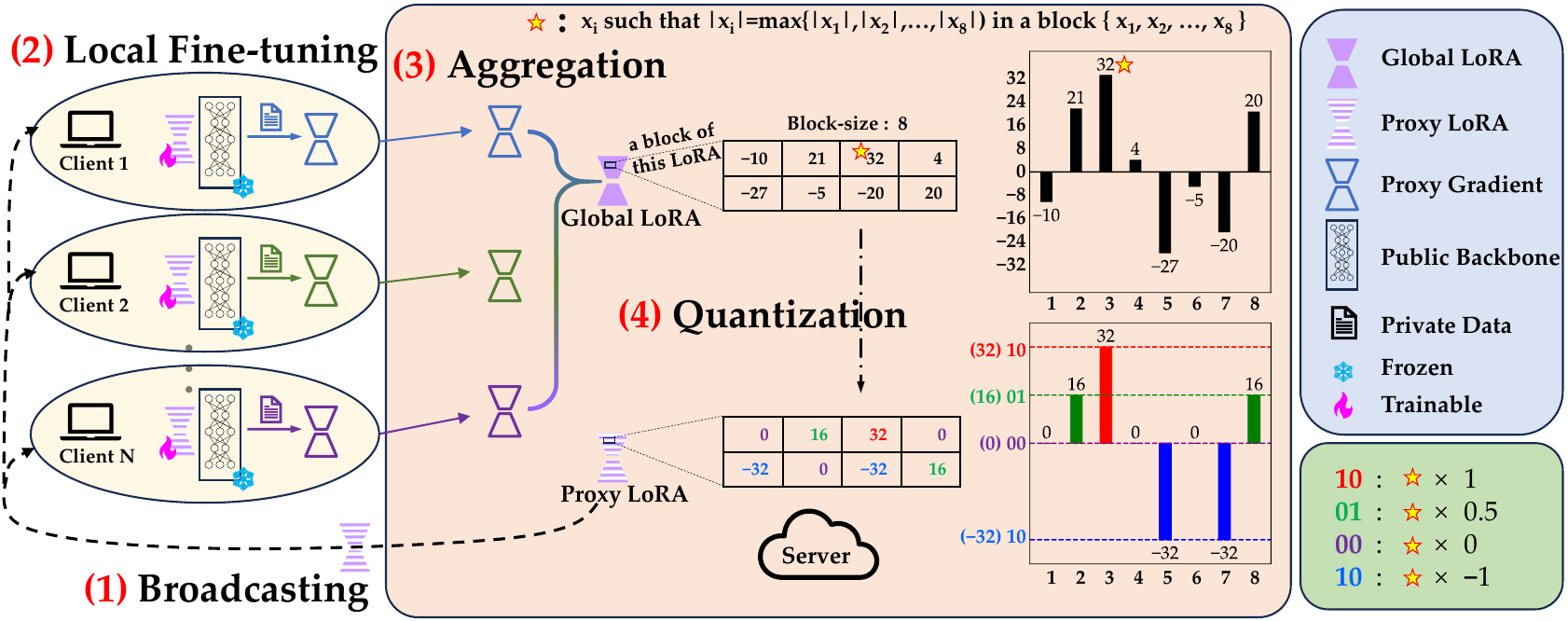}
    \caption{Visualization of the proposed method. To prevent clients (or edge devices) from fully accessing to a global model, the central server broadcasts only the quantized version of the global LoRA's parameters at each communication round. The two bar charts visually demonstrate the quantization process, wherein eight distinct parameter values are mapped into four categories (highlighted in red, green, red, purple and blue). 
     }
    \label{fig:1}
\end{figure*}

\section{Introduction}

As large language models (LLMs) \citep{radford2019language, brown2020language, zhang2022opt, chowdhery2023palm, le2023bloom, zeng2022glm, achiam2023gpt, touvron2023llama} continue to advance, their applications are proliferating across various fields, including healthcare \citep{ge2020fedner}, finance \citep{long2020federated}, and the mobile keyboard \citep{ji2019learning}. These applications often involve training LLMs on data that is distributed across multiple clients or edge devices, with stringent privacy constraints imposed on this data. Federated Learning (FL) \citep{konecny2016federated, mcmahan2017communication, yang2019federated, kairouz2021advances} emerges as a promising paradigm in such scenarios, allowing for the training of models on decentralized data without requiring the transfer of raw data to a central server.

In traditional FL, the primary focus has been on ensuring data privacy. The central server sends the global model to clients, who update the model parameters locally and then send the updates back to the server. However, as LLMs become more sophisticated and valuable, the models themselves become critical intellectual property that also needs protection. This necessity is particularly evident when LLMs are commercialized and provided as paid services, wherein unauthorized access to the models could significantly undermine the interests of the developers and owners.

This new scenario in FL delineates that clients, as custodians of private data, are restricted from accessing the model on the server. Their participation is limited to computational tasks within the FL course. Similarly, the server, as the owner of the FL product, is not authorized to collect data dispersed among different clients. This raises the question: Is there a framework that can protect both data and model privacy simultaneously?

Unfortunately, existing FL frameworks mostly focus on protecting data privacy and have not been dedicated to protecting model privacy, leaving the model's intellectual property vulnerable to potential breaches. Take \textsc{FedAvg}, which is the most widely used FL algorithm, for  example, clients always have access to the global model maintained by the server, which is the one the server intends to commercialize. It will pose a significant risk to the global model's intellectual property and the interests of this model's owners. Hence, there is a crucial need for mechanisms that ensure ``LLM Privacy Protection'', which means preventing clients (or edge devices) from obtaining a model that performs comparably to the final global model during the FL process.

To the best of our knowledge, only one existing study, known as \textsc{FedSP} \citep{dong2023tunable}, has attempted to address LLM privacy protection in FL. The \textsc{FedSP} approach involves constructing a proxy model at the onset of the FL process, which clients download from the server instead of the actual global LLM, thereby circumventing the need to share the global LLM. However, \textsc{FedSP} necessitates the server to have access to a labeled dataset that is identically and independently distributed (iid) with the clients' data, utilizing this dataset for fine-tuning in each communication round. 
This requirement is highly uncommon in real-world FL scenarios.
As the implications of the server's fine-tuning on this dataset, it is unclear whether the model's performance improvements are due to the contributions of the distributed clients or simply the result of the server's fine-tuning on this dataset. 
As demonstrated by our experiments, \textsc{FedSP}'s performance degrades significantly without this fine-tuning on such a dataset, highlighting that its effectiveness heavily relies on the labeled dataset at the server rather than the proposed FL algorithm.

Recognizing these limitations, we introduce a novel approach, \textsc{FedLPP} (FL with LLM Privacy Protection) , to address the dual challenges of protecting both data and model privacy. As shown in Fig \ref{fig:1}, Our proposed framework leverages quantization techniques and parameter-efficient fine-tuning (PEFT) \citep{houlsby2019parameter, lester2021power, li2021prefix, hu2021lora, zaken2021bitfit, wu2024advancing} strategies to ensure that clients can only access a quantized version of the model's parameters. This prevents clients from obtaining a model that performs as well as the centrally hosted one while still enabling accurate gradient estimations for parameter updates. Additionally, by combining this quantization strategy with LoRA \citep{hu2021lora}, we significantly reduce communication costs in FL.

In summary, our framework, \textsc{FedLPP},  effectively ensures both data and model privacy in FL contexts. The learned central model exhibits strong generalization capabilities and can be trained in a resource-efficient manner, addressing the dual privacy challenges in FL. Our contributions are as follows:
\begin{itemize}
    \item \textbf{Privacy Protection for Both Models and Data}. The proposed algorithm enables effective learning without requiring model owners to open-source their models or data owners to share their data. Achieves mutual confidentiality between the server and clients.
    \item \textbf{Excellent Performance}. Experiments on four text generation datasets demonstrate the great performance improvement of our method compared to the baseline. Our approach achieves model privacy protection without significant performance degradation.
    \item \textbf{Low Communication and Computation Demands}. Our method is also applicable in scenarios with limited computational and communication resources, making it well-suited for real-world applications.
\end{itemize}

\section{Related Work}

\subsection{Federated Learning}
Federated learning (FL)\citep{konecny2016federated, mcmahan2017communication, yang2019federated, kairouz2021advances} is a distributed machine learning approach that enables model training across multiple devices or clients while keeping data decentralized and privacy-preserving. FL has gained significant attention due to its ability to address privacy concerns in sensitive data scenarios. Research in FL has primarily focused on overcoming challenges \citep{wen2023survey}such as Privacy and security challenges\citep{bogdanov2008sharemind, geyer2017differentially, cai2020under}, communication challenges\citep{shahid2021communication} and heterogeneity challenges\citep{wang2020towards}. Notable works in FL include \textsc{FedAvg}\citep{mcmahan2017communication}, \textsc{FedProx}\citep{li2020federated}, and \textsc{FedGAN}\citep{rasouli2020fedgan}, with \textsc{FedAvg}\citep{mcmahan2017communication} being a widely recognized algorithm that works by averaging the model updates received from multiple clients, allowing for collaborative model training without sharing raw data.

\subsection{LLMs-Privacy-Protection in Federated Learning}
The previously mentioned \textsc{FedSP}\citep{dong2023tunable} is the only notable study that has attempted to achieves LLM privacy protection in FL, by broadcasting a proxy LLM to clients. This proxy LLM is crucial because clients can only download it from the server, not the actual global LLM, thereby safeguarding the intellectual property of the global LLM and the commercial interests of its owners.

The proxy LLM is created by selecting specific layers of parameters from the global LLM, replicating these layers to match the number of layers of the global LLM, and then aligning the proxy LLM with the global LLM through knowledge distillation (KD)\citep{hinton2015distilling} techniques. This alignment ensures that the proxy model received by clients is structurally similar to the global LLM and performs comparably in terms of model performance.

Next, after obtaining the proxy model, \textsc{FedSP} clients utilize the Prefix Tuning\citep{li2021prefix} technique for training. Since the structure of the proxy model is identical to the global LLM, the soft prompts derived from the clients' training can be applied to the global LLM, facilitating effective information exchange between clients and the server. In each communication round, the server broadcasts a soft prompt to each client, who then trains it on its local private data. The server collects and averages these trained soft prompts to derive an aggregated soft prompt, which is then broadcast to each client at the start of the next communication round.

However, the aggregated soft prompts do not directly align with the global LLM on the server, as clients have inserted them into the proxy model and fine-tuning it. Therefore, \textsc{FedSP} needs the server to insert this aggregated soft prompt into the global LLM and fine-tune on a labeled dataset that is identically-and-independently-distributed (iid) with the client data, to make the aggregated soft prompt applicable to the global LLM. However, the existence of such a dataset contradicts the fundamental principle of FL, which focuses on how the server can learn without collecting data from clients. If an FL algorithm relies on the server "acquiring" a dataset sufficiently similar to the client data so that the server directly fine-tune on it, it will undermine the credibility of the technique and poses a challenge to client data privacy. Therefore, we propose \textsc{FedLPP} to simultaneously ensures both data and model privacy in the federated learning context.

\subsection{Quantization for Federated Learning}

Research on quantization techniques in FL has predominantly focused on reducing communication bandwidth or computational overheads, without addressing the potential application of these technologies to LLMs-Privacy-Protection. A milestone in this field, \textsc{FedPAQ} \citep{reisizadeh2020fedpaq}, introduces a FL framework with quantization, where each client uploads quantized local updates to the server. In contrast, we apply the quantization operation to the LoRA parameters sent by the server to the client.

\subsection{Parameter-Efficient Fine-Tuning for Federated Learning}
Parameter-Efficient Fine-Tuning (PEFT)\citep{houlsby2019parameter, lester2021power, li2021prefix, hu2021lora, zaken2021bitfit, wu2024advancing} has gained significant attention in recent years, with techniques such as Prefix Tuning \citep{li2021prefix} and LoRA \citep{hu2021lora} demonstrating effective strategies for reducing model parameters while maintaining performance. 
LoRA is a notable technique which uses low-rank matrices to update weight matrices during fine-tuning. Instead of directly updating the pre-trained weights, LoRA represents updates as a low-rank decomposition $\Delta W = BA$, where $B$ and $A$ are smaller matrices. This reduces computational complexity and memory usage while maintaining performance.

Leveraging these feature, the PEFT methods can be used to reduce communication overhead and alleviate the training burden on individual clients in FL. In the field of Natural Language Processing (NLP), FedPETuning \citep{zhang2023fedpetuning} provides a benchmark for a comprehensive evaluation of PEFT methods for LLMs under FL settings. Our work also demonstrates that How PEFT methods contribute to LLM privacy protection.

\begin{table*}[t]
\centering
\setlength{\tabcolsep}{9.5pt}
\small
\begin{tabular}{lcrrrcc}
\toprule
\textbf{Dataset} & \textbf{FL Scenario} & \textbf{$\#$Train} & \textbf{$\#$Validation} & \textbf{$\#$Test} & \textbf{N} & \textbf{C} \\
\midrule
\textsc{E2E} & large-scale cross-device & 42,061 & 4,672 & 46,93 & 25 & 5 \\
\textsc{E2E} & cross-silo & 42,061 & 46,72 & 4,693 & 5 & 5 \\
\textsc{ViGGO} & cross-silo & 5,103 & 714 & 1,083 & 5 & 5 \\
\textsc{DART} & large-scale cross-device & 62,659 & 2,768 & 5,097 & 50 & 5 \\
\textsc{DialogSum} & large-scale cross-device & 12,460 & 500 & 1,500 &20 & 5\\
\bottomrule
\end{tabular}
\caption{Statistics of different datasets. The symbol $N$ denotes the total number of clients and $C$ denotes the actual number of clients participating in each communication round. Different values of $C$ are applied to simulate potential scenarios such as unresponsive clients or errors during synchronization in real-world settings.}
\label{tab:datadetail}
\end{table*}

\begin{table*}[t]
\centering
\setlength{\tabcolsep}{5pt}
\small
\begin{tabular}{lccccccc}
\toprule
\textbf{Method} & \textbf{BLEU} & \textbf{NIST} & \textbf{METEOR} & \textbf{ROUGE-L} & \textbf{CIDEr} & \makecell{\textbf{Data} \\\textbf{Privacy}}  & \makecell{\textbf{Model} \\\textbf{Privacy}}\\
\midrule
\textsc{FedLPP} &&&&&&\multirow{3}{*}{\checkmark} & \multirow{3}{*}{\checkmark}\\
\hspace{1em} $\llcorner$ Global& \textbf{34.60} & \textbf{6.06} & \textbf{31.43} & \textbf{51.32} & \textbf{1.70} &  &\\
\hspace{1em} $\llcorner$ Proxy & 32.46 & 5.12 & 29.72 & 50.31 & 1.52& & \\
\midrule
\textsc{FedSP} & 26.42   &  3.65 & 25.88 & 44.42 & 1.21 & \ding{55} & \checkmark \\
\hspace{1em} $\llcorner$(w/o Server Train) & 0.14 & 0.20 & 2.91 & 8.57 & 0.00 & \checkmark & \checkmark \\
\hspace{1em} $\llcorner$(w/o Client Train) & 29.64 & 4.85 & 27.51 & 46.74 & 1.38 & \ding{55} & \checkmark \\
\cmidrule{1-8}
\textsc{FedAvg} + LoRA & 36.04 & 6.58 & 33.64 & 53.01 & 1.90 & \checkmark & \ding{55}\\
\bottomrule
\end{tabular}
\caption{Results achieved by the proposed method and the baselines. For \textsc{FEDLPP}, we use bold formatting to distinguish the better one between Global model and Proxy model. Our proposed method \textsc{FedLPP} has achieved a significant improvement compared to the baseline method \textsc{FedSP}, while simultaneously protecting both the model and the data.}
\label{tab:mainresult}
\end{table*}

\section{Methods}
\subsection{Preliminary}

In a cross-device scenario with $N$ clients, where client $i$
owns a private dataset ${\mathcal D}_i$,  the standard Federated Learning (FL) considers training the weight matrix $\textbf{W}$ by minimizing the loss (empirical risk):

\begin{equation}
	\mathop{\min}_{\textbf W} {\mathcal L}({\textbf W}) = \sum_{n=1}^{N} \frac{n_i}{n}{\mathcal L}_n({\textbf W})
\end{equation}
where  ${\mathcal L}_n(\cdot)$ is the local loss function on ${\mathcal D}_n$, $|{\mathcal D}_i|=n_i$,$n=\sum_{i=1}^{N} n_i$, $\textbf W \in {\mathcal R}^{d \times k}$.

To reduce the communication overhead in the FL course, we use Low-Rank Adaptation technology to decompose $\textbf{W}$ into a frozen pre-trained weight matrix $\textbf{W}_0 \in {\mathcal R}^{d \times k}$ and the trainable delta matrices ${\textbf B} \in {\mathcal R}^{d \times r}$ and ${\textbf A} \in {\mathcal R}^{r \times k}$, where $r$ is the rank of LoRA:
\begin{equation}
{\textbf W} = {\textbf W}_{0} + {\textbf B}{\textbf A},
\end{equation}
Here, $\textbf{W}_0$ is open-source and available online for both the server and the clients, while ${\textbf B}$ and ${\textbf A}$  contain task-specific parameters and enriched with proprietary knowledge after FL course. Therefore, the privacy of ${\textbf B}$ and ${\textbf A}$, which are the final product of the FL course, is crucial for maintaining the commercial interests of their owners, which means the server should avoid directly sharing ${\textbf B}$ and ${\textbf A}$ with the clients, while the server also needs to collect updates of the ${\textbf B}$ and ${\textbf A}$ matrices from the clients, enabling effective FL course:
\begin{equation}
{\textbf X}^{t+1} = {\textbf X}^{t} + \sum_{i=1}^{N}\frac{n_i}{n} \Delta {\textbf X}_i^{t}
\label{equ:3}
\end{equation}
In Equation \ref{equ:3}, ${\textbf X}$ can represent either ${\textbf B}$ or ${\textbf A}$, depending on the context, and $t \in \{1,2,...,T\}$ means the $t$-th communication round of the FL course, T is the total number of communication rounds.

\subsection{Computing Proxy LoRA Matrices by Quantization}

In traditional deep neural networks (DNNs), full-precision model parameters are typically stored in 32-bit floating-point format. To preserve the privacy of the global LoRA matrices ${\textbf X} \in  {\mathcal R}^{a \times b}$, \textsc{FedLPP} no longer sends the exact values of ${\textbf X}$ to the clients. Instead, the server computes low-precision versions ${\mathbf{\textbf X}} = Q({\textbf X})$ as proxy LoRA matrices which are available to clients. The specific calculation process of $Q(\cdot)$ is as follows.

First, following \textsc{QLoRA}\citep{dettmers2024qlora}, after selecting the desired bit-width $w$ for quantization, we then select $2^{(w-1)}-2$ numbers from $(-1,0)$ and $2^{(w-1)}-1$ numbers from $(0,1)$, combine them with $-1$, $0$, and $1$, totaling $2^w$ numbers. We refer to these numbers as "standard numbers" and denote them in sorted order as $V = \{v_0,v_1,...,v_{2^w-1}\}$, where $v_0=-1$, $v_{2^{(w-1)}-1}=0$, and $v_{2^w-1}=1$. 

Second, Combine consecutive floating-point numbers in matrix ${\textbf X}$ into blocks of size $s$, maintaining the spatial continuity\footnote{Here, 'spatial continuity' refers to the uninterrupted sequence of floating-points numbers in the flattened representation of matrix ${\textbf X}$.} of ${\textbf X}$. Identify the floating-point number with the largest absolute value within the $l$-th block $X_l$ as $z_l$, and form an array $Z$ of length $\lfloor \frac{ab}{s} \rfloor$ using these $z_l$ values.Therefore, we can get the normalized version of the $l$-th block $X_l$ as:
\begin{equation}
{X'}_{l} = \frac{{X}_{l}}{z_l}
\end{equation}

Thus, the floating-point numbers in ${X'}_l$ fall within the same value range as the "standard numbers", So we can replace all floating-point numbers in ${X'}_l$ with the closest "standard numbers" to obtain a block of the same size as ${X'}_l$, denoted as ${\tilde{X'}}_l$, which is an approximate version of ${X'}_l$. Therefore, ${\tilde{X}}_l = z_l{\tilde{X'}}_l$ is also an approximate version of ${X}^l$, as ${X}^l = z_l{X'}_{l}$. When we arrange all $\lfloor \frac{ab}{s} \rfloor$ blocks $\tilde{X}_l$ in their original order to form $\tilde{\textbf{X}}$, we obtain an approximate version of the original $\mathbf{X}$.

It is noted that the computed $\tilde{\textbf{X}}$ compared to $\mathbf{X}$ incurs some information loss, if we broadcast $\tilde{\textbf{X}}$ as a proxy LoRA matrix to the clients instead of $\mathbf{X}$, clients cannot get a model with comparable performance to the one owned by the server. However, since $0 \in V$, we know that the transformation from $\mathbf{X}$ to $\tilde{\textbf{X}}$ preserves the sign of all floating-point numbers in $\mathbf{X}$. Consequently, the proxy gradients computed by clients using this proxy LoRA matrix will serve as a good estimate of the true gradients. Hence, our constructed proxy LoRA matrix holds the potential to achieve both data privacy protection and LLM privacy protection simultaneously.

\subsection{Client-Side Local Fine-tuning}

Upon receiving the trainable proxy LoRA matrix$\tilde{\textbf{X}}^t$  in the $t$-th round, the selected client $i$ assembles them with the frozen backbone ${\textbf W}_0$ to form a local model. Subsequently, using ${\mathcal L}_i$ and local private data ${\mathcal D}_i$, the local model undergoes training to obtain proxy update $\Delta {\tilde{\textbf X}}_i^t$, which are then collected by the server. The central server employs secure aggregation algorithms\citep{bonawitz2016practical} to compute ${\textbf X^{t+1}}$ for the next round. This process iterates continuously until the communication round $t$ reaches the upper limit $T$.

\begin{table*}[t]
\centering
\setlength{\tabcolsep}{8pt}
\small
\begin{tabular}{lccccccc}
\toprule
\textbf{Method} & \textbf{BLEU} & \textbf{NIST} & \textbf{METEOR} & \textbf{ROUGE-L} & \textbf{CIDEr}  & \makecell{\textbf{Data} \\\textbf{Privacy}}  & \makecell{\textbf{Model} \\\textbf{Privacy}} \\
\midrule
\textsc{FedLPP} ($w=1$)&&&&&&\multirow{3}{*}{\checkmark} & \multirow{3}{*}{\checkmark}\\
\hspace{1em} $\llcorner$ Global & \textbf{51.81} & \textbf{6.66} & \textbf{0.33} & \textbf{60.30} & \textbf{1.29}  &  & \\
\hspace{1em} $\llcorner$ Proxy & 51.54 & 6.15 & 0.33 & 59.94 & 1.24 & &\\
\cmidrule{1-8}
\textsc{FedLPP} ($w=2$)&&&&&&\multirow{3}{*}{\checkmark} & \multirow{3}{*}{\checkmark}\\
\hspace{1em} $\llcorner$ Global & \textbf{54.93} & \textbf{7.70} & \textbf{0.37} & \textbf{61.65} & \textbf{1.40} &  & \\
\hspace{1em} $\llcorner$ Proxy & 52.05 & 6.23 & 0.35 & 60.75 & 1.22  & &\\
\cmidrule{1-8}
\textsc{FedLPP} ($w=3$)&&&&&&\multirow{3}{*}{\checkmark} & \multirow{3}{*}{\ding{55}}\\
\hspace{1em} $\llcorner$ Global & 54.62 & 7.71 & \textbf{0.41} & \textbf{62.33} & \textbf{1.45}  & &\\
\hspace{1em} $\llcorner$ Proxy & \textbf{54.97} &\textbf{7.74} & 0.38 & 61.65 & 1.42  & &\\
\cmidrule{1-8}
\textsc{FedAvg} + LoRA & 54.88 & 7.71 & 0.41 & 63.24 & 1.50 &\checkmark  &\ding{55}\\
\bottomrule
\end{tabular}
\caption{Results achieved by the proposed method using varying quantization levels.With an appropriate quantization level, specifically when $w$=2, \textsc{FedLPP} can achieve simultaneous model and data protection while maintaining high performance.}
\label{tab:bwresult}
\end{table*}
\begin{table*}[t]
\centering
\setlength{\tabcolsep}{4.5pt}
\small
\begin{tabular}{clccccccc}
\toprule
\textbf{FL Scenario} & \textbf{Method} & \textbf{BLEU} & \textbf{NIST} & \textbf{METEOR} & \textbf{ROUGE-L} & \textbf{CIDEr} & \makecell{\textbf{Data} \\\textbf{Privacy}}  & \makecell{\textbf{Model} \\\textbf{Privacy}} \\
\midrule
\multirow{4}{*}{\textbf{Cross-Silo}}&\textsc{FedLPP} &&&&&&\multirow{3}{*}{\checkmark} & \multirow{3}{*}{\checkmark}\\
&\hspace{1em} $\llcorner$ Global & \textbf{54.75} & \textbf{7.68} & \textbf{0.41} & \textbf{62.66} & \textbf{1.47} &  &\\
&\hspace{1em} $\llcorner$ Proxy & 53.42 & 7.56 & 0.38 & 61.12 & 1.36 & &\\
\cmidrule{2-9}
&\textsc{FedAvg}+LoRA & 54.99 & 7.73 & 0.41 & 63.32 & 1.51 &\checkmark  &\ding{55}\\
\cmidrule{1-9}
\multirow{4}{*}{\makecell{\textbf{Large-Scale} \\\textbf{Cross-Device}}}&\textsc{FedLPP} &&&&&&\multirow{3}{*}{\checkmark} & \multirow{3}{*}{\checkmark}\\
&\hspace{1em} $\llcorner$ Global & \textbf{54.93} & \textbf{7.70} & \textbf{0.37} & \textbf{61.65} & \textbf{1.40} &  & \\
&\hspace{1em} $\llcorner$ Proxy& 52.05 & 6.23 & 0.35 & 60.75 & 1.22  & &  \\
\cmidrule{2-9}
&\textsc{FedAvg}+LoRA & 54.88 & 7.71 & 0.41 & 63.24 & 1.50 &\checkmark  &\ding{55} \\
\bottomrule
\end{tabular}
\caption{Results achieved by the proposed method under different FL scenario. After transitioning our FL Scenario from Cross-Silo to Large-Scale Cross-Device, FedLPP's performance has not been significantly impacted and still fulfills the task of protecting the model.}
\label{tab:dataresult}
\end{table*}

\begin{table*}[t]
\centering
\setlength{\tabcolsep}{6pt}
\small
\begin{tabular}{llccccccc}
\toprule
\textbf{Model}&\textbf{Method} & \textbf{BLEU} & \textbf{NIST} & \textbf{METEOR} & \textbf{ROUGE-L} & \textbf{CIDEr} & \makecell{\textbf{Data} \\\textbf{Privacy}}  & \makecell{\textbf{Model} \\\textbf{Privacy}} \\
\midrule
\multirow{4}{*}{\makecell{\textbf{GPT-2} \\\textbf{Medium}}}& \textsc{FedLPP} & &  &  & &  & \multirow{3}{*}{\checkmark} &\multirow{3}{*}{\checkmark} \\
& \hspace{1em} $\llcorner$ Global & \textbf{54.93} & \textbf{7.70} & \textbf{0.37} & \textbf{61.65} & \textbf{1.40} &  & \\
&\hspace{1em} $\llcorner$ Proxy & 52.05 & 6.23 & 0.35 & 60.75 & 1.22  & & \\
\cmidrule{2-9}
&\textsc{FedAvg}+LoRA & 54.88 & 7.71 & 0.41 & 63.24 & 1.50 &\checkmark  &\ding{55}\\
\cmidrule{1-9}
\multirow{4}{*}{\makecell{\textbf{GPT-2} \\\textbf{XL}}}&\textsc{FedLPP} & &  &  &  &  & \multirow{3}{*}{\checkmark} &\multirow{3}{*}{\checkmark}\\
&\hspace{1em} $\llcorner$ Global & \textbf{54.81}& \textbf{7.76} & \textbf{0.39} & \textbf{61.34} & \textbf{1.41} &  &\\
&\hspace{1em} $\llcorner$ Proxy & 51.58 & 7.03 & 0.35 & 59.51 & 1.23 &  &\\
\cmidrule{2-9}
&\textsc{FedAvg}+LoRA &  55.52 & 7.77 & 0.41 & 63.01 & 1.50 &\checkmark  &\ding{55}\\
\bottomrule
\end{tabular}
\caption{Results achieved by the proposed method with models of varying size.}
\label{tab:scalresult}
\end{table*}

\section{Experiments}

In order to test whether our algorithm maintains good FL performance while simultaneously preserving the privacy of both the model and the data, extensive experiments are conducted in this section, including performance comparisons (see Sec \ref{sspc}) . Additionally, ablation analysis of \textsc{FedLPP} was also conducted, including different FL scenarios (see Sec \ref{ssdfc}) , different quantization level (see Sec \ref{ssdql}) and scaling (see Sec \ref{ssdfc}) .

\subsection{Models and Datasets}
\label{sses}
We conduct our experiments with two popular Language models, i.e., GPT2-XL and GPT2-Medium\citep{radford2019language}, and our experiments utilized 4 datasets: \textsc{E2E} \citep{novikova2017e2e}, \textsc{ViGGO} \citep{juraska2019viggo}, \textsc{DART} \citep{nan2020dart} and \textsc{DialogSUM} \citep{chen2021dialogsum}.
The \textsc{E2E} dataset comprises a collection of tabel-to-text generation data for training end-to-end natural language generation systems in the restaurant domain. \textsc{ViGGO} dataset is also a tabel-to-text generation dataset but was designed for generalizable and conversational dialogue act types. \textsc{DART} is another open-domain tabel-to-text generation dataset and \textsc{DialogSUM} is a dataset designed for dialogue summarization. We use 5 metrics to evaluate the quality of the text generated by our model: BLEU\citep{papineni2002bleu}, NIST\citep{belz2006comparing}, METEOR\citep{banerjee2005meteor}, ROUGE-L\citep{lin2004rouge}, CIDEr\citep{vedantam2015cider}. More detials of how we use these 4 datasets for FL setting is shown in Table \ref{tab:datadetail}.

\subsection{Baselines}
\label{ssb}
We use the following four methods as baselines to compare with our proposed FedLPP:
\begin{itemize}
\item \textbf{\textsc{FedSP}}\citep{dong2023tunable}: This is the only framework that focuses on addressing the issue of LLM privacy protection in FL up to date.
\item \textbf{\textsc{FedSP} (w/o Server Train)}: A variant of FedSP where the server does not use a labeled dataset that is iid with the client's data for supervised training. However, FedSP can still utilize the unlabeled version of this dataset for the Knowledge Distillation (KD) process to create a proxy model at the start of its FL process. This variant helps verify whether FedSP's performance stems from the server's labeled dataset or the private datasets on other participating clients in the FL course.
\item \textbf{\textsc{FedSP} (w/o Client Train)}: Another variant of FedSP where clients do not join in the FL, Which means that the server just trains the model in its labeled dataset. These two variants help verify whether FedSP's performance stems from the server's labeled dataset or the private datasets of other participating clients in the FL course.
\item \textbf{\textsc{FedAVG}$+$LoRA}: Uses the \textsc{FedAVG} algorithm to train LoRA adapters while keeping the backbone frozen. It is also a variant of our method which doesn't quantify the LoRA matrics before broadcast it to the clients. Since this method does not consider LLM privacy protection (i.e., clients can have unrestricted access to the server's LoRA adapters), we only consider it as a potential upper bound for comparison with the \textsc{FedLPP} algorithm.
\end{itemize}
\subsection{Implementation Details}
\label{ssid}

For fair and reasonable comparisons, we conducted hyper-parameter searching for each dataset and method. We selected the best model based on loss from the validation set and reported metrics on the test set. 

We primarily evaluated the performance of \textsc{FedLPP} under three quantization levels, namely bit-width $ w \in \{1,2,3\}$, with the same block-size 256. As for FedSP, the prefix length was chosen from $\{40, 80, 160\}$, and the layer number of the proxy model was chosen from $\{1, 4, 8\}$. For all methods, the learning rate was chosen from $\{1e-4, 3e-4, 1e-3\}$, and the batch size is set to 16. The training epochs in each round were chosen from $\{1, 3, 5\}$ and the total number of communication rounds is set to 100. We implement the proposed approach FedLPP and baseline methods based on Huggingface Transformers\citep{wolf2020transformers}. All experiments were conducted on a single server equipped with 4 NVIDIA GeForce RTX 3090 GPUs, each with 24GB of RAM.

\subsection{Performance Comparison}

\label{sspc}
\begin{figure*}[ht]
\centering    \includegraphics[width=0.95\linewidth]{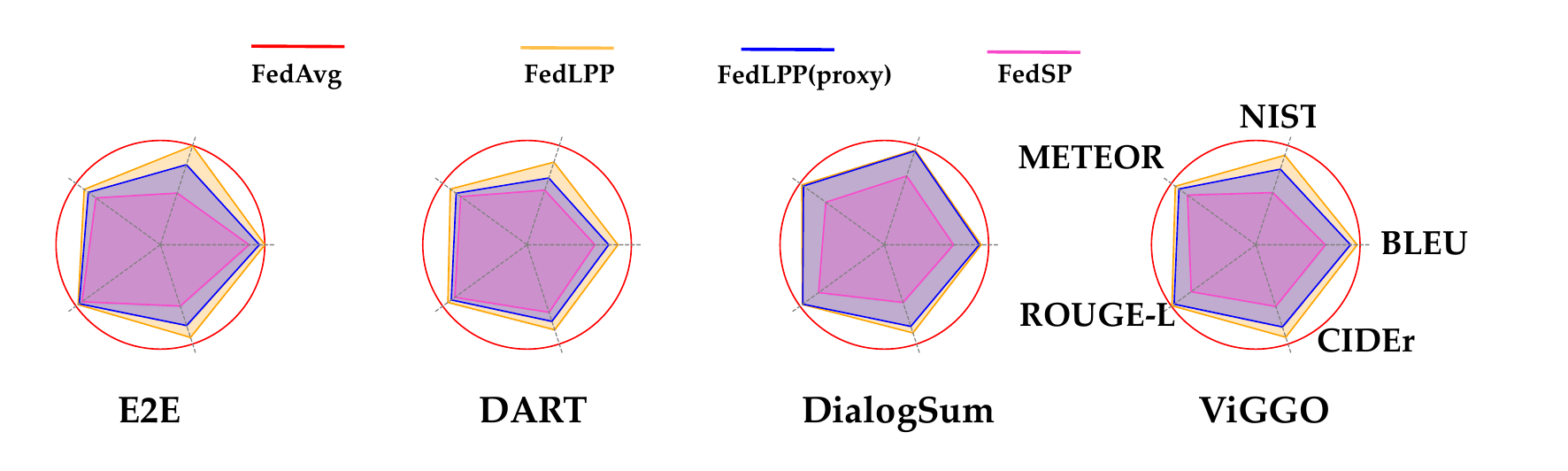}
\caption{Performance comparison of various methods on four different datasets. As shown in the radar chart, On all four datasets, \textsc{FedLPP} has achieved the goal of protecting the model and demonstrated significant performance improvements compared to the baseline \textsc{FedSP}.
 }
\label{fig:2}
\vspace{-6mm}
\end{figure*}

In Table \ref{tab:mainresult}, we present comparisons between our method and the baselines. For all of the methods, we select the best models among all communication rounds as the final model of the FL course. For our approach, we report both the best global model on the server and the best proxy model accessible to clients. We only consider \textsc{FedLPP} to have effectively protected the global model when the global model outperforms the proxy model in all metrics. Table \ref{tab:mainresult} shows the average performance across the four datasets, and the performance for each dataset is shown in the radar charts in Figure \ref{fig:2}.

Based on Table \ref{tab:mainresult} and Figure \ref{fig:2}, we can draw the following conclusions. Our method shows significant improvements over \textsc{FedSP} across four datasets and maintains comparable performance to \textsc{FedAvg}+LoRA. As \textsc{FedAvg}+LoRA can be seen as a variant of \textsc{FedLPP} without model protection, we can infer from the results that our additional efforts to protect the model did not lead to a significant decrease in performance.

Additionally, we compare the global model with the proxy model in the \textsc{FedLPP} algorithm. Analyzing the first two rows in Table \ref{tab:mainresult} , we find that the global model on the server consistently outperforms the proxy model across five metrics, ensuring that \textsc{FedLPP} indeed achieves the goal of protecting the commercial interests of the owners of large models. As long as the best model is not disclosed to any participating client, the model owned by the server will consistently maintain an advantage over each client, thus protecting the fundamental interests of the model owner.

Regarding data privacy protection, it is noted that our method is compatible with secure aggregation algorithms\citep{bonawitz2016practical} and we do not need a labeled dataset in the server which will compromise client data privacy, unlike \textsc{FedSP}. Therefore, our method offers security comparable to \textsc{FedAVG} in this regard. 

However, the analysis of the baseline method, \textsc{FedSP},  and its two variants confirms that the performance of \textsc{FedSP} mainly arises from fine-tuning on the server's dataset. Without referencing the aggregated information from clients, the \textsc{FedSP} server can train models with performance even better than the original \textsc{FedSP}, but when FedSP no longer fine-tunes on such a dataset on the server, the algorithm cannot proceed. So the significance of this server dataset to \textsc{FedSP} is apparent. However, as discussed earlier, such an labeled datasets do not meet the requirements of data privacy protection.

\subsection{Impact of Different Quantization Level}
\label{ssdql}
In the FedLPP framework, there is a trade-off between LLM privacy protection and performance. This trade-off depends on the level of quantization: a higher level of quantization results in a greater deviation of the proxy LoRA received by the clients from the global LoRA, leading to increased protection but also a larger bias in the computed proxy gradients, which impacts the final performance. We can control the level of quantization by adjusting bit-width $w$ and the block size $s$, and even generalize this approach without limiting the length of V to a power of 2 for more nuanced trade-offs. In this study, we experimented with three different quantization bit widths: $\{1, 2, 3\}$, and the results are shown in the Table \ref{tab:bwresult}. It is evident that choosing a bit level of 2 achieves better performance while ensuring LLM privacy protection. 

If the quantization level is too low, the information loss in the proxy model received by the client will be insufficient to ensure model privacy protection. This might even result in merely removing noise in the global model. As shown in the table, when $w$=3, the proxy model's performance even surpasses that of the global model in some of the metrics, this implies that the server cannot guarantee the privacy of the model.

\subsection{Impact of Different FL Scenarios and Model Scaling}
\label{ssdfc}
To validate the effectiveness of \textsc{FedLPP} in different FL scenarios, we primarily considered two FL scenarios: cross-silo FL scenario \citep{kairouz2021advances} and large-scale cross-device FL scenario \citep{lai2022fedscale}. In the cross-silo scenario, the server selects all clients for training in each communication round. but in large-scale cross-device scenarios, the data held by local clients is scarcer, additionally, not every client participates in each learning round. This  more closely reflects real-world FL scenario where, due to communication constraints or synchronization issues, the server cannot always receive responses from every client. Therefore, it is necessary to test whether the FedLPP algorithm can perform effectively under these more challenging conditions. In this experiment, we selected \textsc{ViGGO} to simulate cross-silo FL scenarios, and \textsc{Dart}, \textsc{DialogSum} to simulate large-scale cross-device FL scenarios. And we also use \textsc{E2E} to simultaneously simulate these two scenarios to make a clear comparison and study whether the effectiveness of \textsc{FedLPP} is affected by the FL Scenario factor.The performance comparison of \textsc{FedLPP} under two different data splits is shown in Table \ref{tab:dataresult}. 

On the other hand, we also extended the model we use from GPT2-Medium to a larger model, GPT2-XL. To verify whether FedLPP can be extended for use with larger models. The results on the E2E dataset are shown in the Table \ref{tab:scalresult}.

\section{Conclusion}

In this study, we presented \textsc{FedLPP}, a novel federated learning framework designed to address the dual challenges of data privacy and model privacy.
By integrating a quantization strategy with LoRA, \textsc{FedLPP} achieves effective model parameter updates while substantially reducing communication overhead.
Our framework ensures that each client participates in the learning process without accessing a full-performance model, thereby safeguarding the intellectual property of the model developers.
The results demonstrate that \textsc{FedLPP} not only maintains robust privacy protections but also supports the development of a generalized model with reduced resource consumption.
Moreover, \textsc{FedLPP} is particularly suitable for scenarios where data privacy and intellectual property rights are paramount, providing a practical solution in the evolving landscape of federated learning.

\section*{Limitations}

While \textsc{FedLPP} introduces significant advancements in FL, several limitations warrant further investigation.
First, the quantization process, although effective in reducing the model size and protecting intellectual property, may introduce quantization noise, potentially affecting the learning accuracy and convergence rate.
Future work could explore adaptive quantization techniques to mitigate this issue.
Secondly, the integration of LoRA is primarily tested under controlled conditions; its efficacy across diverse network architectures and heterogeneous data distributions remains to be fully evaluated.
Additionally, the computational overhead associated with the implementation of LoRA and quantization in resource-constrained environments needs thorough assessment.
Addressing these limitations could enhance the applicability of \textsc{FedLPP} across a broader range of FL scenarios and contribute to its adoption in industry-standard practices.

\section*{Reproducibility Statement}
The authors have diligently worked to guarantee the reproducibility of the empirical findings presented in this paper. To facilitate reproducibility, the source code for the proposed method has been submitted alongside the paper, and we intend to make the source code publicly available on GitHub upon acceptance. 



\bibliography{custom}

\begin{thebibliography}{46}
\providecommand{\natexlab}[1]{#1}

\bibitem[{Achiam et~al.(2023)Achiam, Adler, Agarwal, Ahmad, Akkaya, Aleman, Almeida, Altenschmidt, Altman, Anadkat et~al.}]{achiam2023gpt}
Josh Achiam, Steven Adler, Sandhini Agarwal, Lama Ahmad, Ilge Akkaya, Florencia~Leoni Aleman, Diogo Almeida, Janko Altenschmidt, Sam Altman, Shyamal Anadkat, et~al. 2023.
\newblock Gpt-4 technical report.
\newblock \emph{arXiv preprint arXiv:2303.08774}.

\bibitem[{Banerjee and Lavie(2005)}]{banerjee2005meteor}
Satanjeev Banerjee and Alon Lavie. 2005.
\newblock Meteor: An automatic metric for mt evaluation with improved correlation with human judgments.
\newblock In \emph{Proceedings of the acl workshop on intrinsic and extrinsic evaluation measures for machine translation and/or summarization}, pages 65--72.

\bibitem[{Belz and Reiter(2006)}]{belz2006comparing}
Anja Belz and Ehud Reiter. 2006.
\newblock Comparing automatic and human evaluation of nlg systems.
\newblock In \emph{11th conference of the european chapter of the association for computational linguistics}, pages 313--320.

\bibitem[{Bogdanov et~al.(2008)Bogdanov, Laur, and Willemson}]{bogdanov2008sharemind}
Dan Bogdanov, Sven Laur, and Jan Willemson. 2008.
\newblock Sharemind: A framework for fast privacy-preserving computations.
\newblock In \emph{Computer Security-ESORICS 2008: 13th European Symposium on Research in Computer Security, M{\'a}laga, Spain, October 6-8, 2008. Proceedings 13}, pages 192--206. Springer.

\bibitem[{Bonawitz et~al.(2016)Bonawitz, Ivanov, Kreuter, Marcedone, McMahan, Patel, Ramage, Segal, and Seth}]{bonawitz2016practical}
Keith Bonawitz, Vladimir Ivanov, Ben Kreuter, Antonio Marcedone, H~Brendan McMahan, Sarvar Patel, Daniel Ramage, Aaron Segal, and Karn Seth. 2016.
\newblock Practical secure aggregation for federated learning on user-held data.
\newblock \emph{arXiv preprint arXiv:1611.04482}.

\bibitem[{Brown et~al.(2020)Brown, Mann, Ryder, Subbiah, Kaplan, Dhariwal, Neelakantan, Shyam, Sastry, Askell et~al.}]{brown2020language}
Tom Brown, Benjamin Mann, Nick Ryder, Melanie Subbiah, Jared~D Kaplan, Prafulla Dhariwal, Arvind Neelakantan, Pranav Shyam, Girish Sastry, Amanda Askell, et~al. 2020.
\newblock Language models are few-shot learners.
\newblock \emph{Advances in neural information processing systems}, 33:1877--1901.

\bibitem[{Cai et~al.(2020)Cai, Niu, Geng, Zhang, Cui, Li, and Chen}]{cai2020under}
Xingjuan Cai, Yun Niu, Shaojin Geng, Jiangjiang Zhang, Zhihua Cui, Jianwei Li, and Jinjun Chen. 2020.
\newblock An under-sampled software defect prediction method based on hybrid multi-objective cuckoo search.
\newblock \emph{Concurrency and Computation: Practice and Experience}, 32(5):e5478.

\bibitem[{Chen et~al.(2021)Chen, Liu, Chen, and Zhang}]{chen2021dialogsum}
Yulong Chen, Yang Liu, Liang Chen, and Yue Zhang. 2021.
\newblock Dialogsum: A real-life scenario dialogue summarization dataset.
\newblock \emph{arXiv preprint arXiv:2105.06762}.

\bibitem[{Chowdhery et~al.(2023)Chowdhery, Narang, Devlin, Bosma, Mishra, Roberts, Barham, Chung, Sutton, Gehrmann et~al.}]{chowdhery2023palm}
Aakanksha Chowdhery, Sharan Narang, Jacob Devlin, Maarten Bosma, Gaurav Mishra, Adam Roberts, Paul Barham, Hyung~Won Chung, Charles Sutton, Sebastian Gehrmann, et~al. 2023.
\newblock Palm: Scaling language modeling with pathways.
\newblock \emph{Journal of Machine Learning Research}, 24(240):1--113.

\bibitem[{Dettmers et~al.(2024)Dettmers, Pagnoni, Holtzman, and Zettlemoyer}]{dettmers2024qlora}
Tim Dettmers, Artidoro Pagnoni, Ari Holtzman, and Luke Zettlemoyer. 2024.
\newblock Qlora: Efficient finetuning of quantized llms.
\newblock \emph{Advances in Neural Information Processing Systems}, 36.

\bibitem[{Dong et~al.(2023)Dong, Xie, Ding, Shen, and Li}]{dong2023tunable}
Chenhe Dong, Yuexiang Xie, Bolin Ding, Ying Shen, and Yaliang Li. 2023.
\newblock Tunable soft prompts are messengers in federated learning.
\newblock \emph{arXiv preprint arXiv:2311.06805}.

\bibitem[{Ge et~al.(2020)Ge, Wu, Wu, Qi, Huang, and Xie}]{ge2020fedner}
Suyu Ge, Fangzhao Wu, Chuhan Wu, Tao Qi, Yongfeng Huang, and Xing Xie. 2020.
\newblock Fedner: Privacy-preserving medical named entity recognition with federated learning.
\newblock \emph{arXiv preprint arXiv:2003.09288}.

\bibitem[{Geyer et~al.(2017)Geyer, Klein, and Nabi}]{geyer2017differentially}
Robin~C Geyer, Tassilo Klein, and Moin Nabi. 2017.
\newblock Differentially private federated learning: A client level perspective.
\newblock \emph{arXiv preprint arXiv:1712.07557}.

\bibitem[{Hinton et~al.(2015)Hinton, Vinyals, and Dean}]{hinton2015distilling}
Geoffrey Hinton, Oriol Vinyals, and Jeff Dean. 2015.
\newblock Distilling the knowledge in a neural network.
\newblock \emph{arXiv preprint arXiv:1503.02531}.

\bibitem[{Houlsby et~al.(2019)Houlsby, Giurgiu, Jastrzebski, Morrone, De~Laroussilhe, Gesmundo, Attariyan, and Gelly}]{houlsby2019parameter}
Neil Houlsby, Andrei Giurgiu, Stanislaw Jastrzebski, Bruna Morrone, Quentin De~Laroussilhe, Andrea Gesmundo, Mona Attariyan, and Sylvain Gelly. 2019.
\newblock Parameter-efficient transfer learning for nlp.
\newblock In \emph{International conference on machine learning}, pages 2790--2799. PMLR.

\bibitem[{Hu et~al.(2021)Hu, Shen, Wallis, Allen-Zhu, Li, Wang, Wang, and Chen}]{hu2021lora}
Edward~J Hu, Yelong Shen, Phillip Wallis, Zeyuan Allen-Zhu, Yuanzhi Li, Shean Wang, Lu~Wang, and Weizhu Chen. 2021.
\newblock Lora: Low-rank adaptation of large language models.
\newblock \emph{arXiv preprint arXiv:2106.09685}.

\bibitem[{Ji et~al.(2019)Ji, Pan, Long, Li, Jiang, and Huang}]{ji2019learning}
Shaoxiong Ji, Shirui Pan, Guodong Long, Xue Li, Jing Jiang, and Zi~Huang. 2019.
\newblock Learning private neural language modeling with attentive aggregation.
\newblock In \emph{2019 International joint conference on neural networks (IJCNN)}, pages 1--8. IEEE.

\bibitem[{Juraska et~al.(2019)Juraska, Bowden, and Walker}]{juraska2019viggo}
Juraj Juraska, Kevin~K Bowden, and Marilyn Walker. 2019.
\newblock Viggo: A video game corpus for data-to-text generation in open-domain conversation.
\newblock \emph{arXiv preprint arXiv:1910.12129}.

\bibitem[{Kairouz et~al.(2021)Kairouz, McMahan, Avent, Bellet, Bennis, Bhagoji, Bonawitz, Charles, Cormode, Cummings et~al.}]{kairouz2021advances}
Peter Kairouz, H~Brendan McMahan, Brendan Avent, Aur{\'e}lien Bellet, Mehdi Bennis, Arjun~Nitin Bhagoji, Kallista Bonawitz, Zachary Charles, Graham Cormode, Rachel Cummings, et~al. 2021.
\newblock Advances and open problems in federated learning.
\newblock \emph{Foundations and trends{\textregistered} in machine learning}, 14(1--2):1--210.

\bibitem[{Konecn{\`y} et~al.(2016)Konecn{\`y}, McMahan, Yu, Richt{\'a}rik, Suresh, and Bacon}]{konecny2016federated}
Jakub Konecn{\`y}, H~Brendan McMahan, Felix~X Yu, Peter Richt{\'a}rik, Ananda~Theertha Suresh, and Dave Bacon. 2016.
\newblock Federated learning: Strategies for improving communication efficiency.
\newblock \emph{arXiv preprint arXiv:1610.05492}, 8.

\bibitem[{Lai et~al.(2022)Lai, Dai, Singapuram, Liu, Zhu, Madhyastha, and Chowdhury}]{lai2022fedscale}
Fan Lai, Yinwei Dai, Sanjay Singapuram, Jiachen Liu, Xiangfeng Zhu, Harsha Madhyastha, and Mosharaf Chowdhury. 2022.
\newblock Fedscale: Benchmarking model and system performance of federated learning at scale.
\newblock In \emph{International conference on machine learning}, pages 11814--11827. PMLR.

\bibitem[{Lester et~al.(2021)Lester, Al-Rfou, and Constant}]{lester2021power}
Brian Lester, Rami Al-Rfou, and Noah Constant. 2021.
\newblock The power of scale for parameter-efficient prompt tuning.
\newblock \emph{arXiv preprint arXiv:2104.08691}.

\bibitem[{Li et~al.(2020)Li, Sahu, Zaheer, Sanjabi, Talwalkar, and Smith}]{li2020federated}
Tian Li, Anit~Kumar Sahu, Manzil Zaheer, Maziar Sanjabi, Ameet Talwalkar, and Virginia Smith. 2020.
\newblock Federated optimization in heterogeneous networks.
\newblock \emph{Proceedings of Machine learning and systems}, 2:429--450.

\bibitem[{Li and Liang(2021)}]{li2021prefix}
Xiang~Lisa Li and Percy Liang. 2021.
\newblock Prefix-tuning: Optimizing continuous prompts for generation.
\newblock \emph{arXiv preprint arXiv:2101.00190}.

\bibitem[{Lin(2004)}]{lin2004rouge}
Chin-Yew Lin. 2004.
\newblock Rouge: A package for automatic evaluation of summaries.
\newblock In \emph{Text summarization branches out}, pages 74--81.

\bibitem[{Long et~al.(2020)Long, Tan, Jiang, and Zhang}]{long2020federated}
Guodong Long, Yue Tan, Jing Jiang, and Chengqi Zhang. 2020.
\newblock Federated learning for open banking.
\newblock In \emph{Federated Learning: Privacy and Incentive}, pages 240--254. Springer.

\bibitem[{McMahan et~al.(2017)McMahan, Moore, Ramage, Hampson, and y~Arcas}]{mcmahan2017communication}
Brendan McMahan, Eider Moore, Daniel Ramage, Seth Hampson, and Blaise~Aguera y~Arcas. 2017.
\newblock Communication-efficient learning of deep networks from decentralized data.
\newblock In \emph{Artificial intelligence and statistics}, pages 1273--1282. PMLR.

\bibitem[{Nan et~al.(2020)Nan, Radev, Zhang, Rau, Sivaprasad, Hsieh, Tang, Vyas, Verma, Krishna et~al.}]{nan2020dart}
Linyong Nan, Dragomir Radev, Rui Zhang, Amrit Rau, Abhinand Sivaprasad, Chiachun Hsieh, Xiangru Tang, Aadit Vyas, Neha Verma, Pranav Krishna, et~al. 2020.
\newblock Dart: Open-domain structured data record to text generation.
\newblock \emph{arXiv preprint arXiv:2007.02871}.

\bibitem[{Novikova et~al.(2017)Novikova, Du{\v{s}}ek, and Rieser}]{novikova2017e2e}
Jekaterina Novikova, Ond{\v{r}}ej Du{\v{s}}ek, and Verena Rieser. 2017.
\newblock The e2e dataset: New challenges for end-to-end generation.
\newblock \emph{arXiv preprint arXiv:1706.09254}.

\bibitem[{Papineni et~al.(2002)Papineni, Roukos, Ward, and Zhu}]{papineni2002bleu}
Kishore Papineni, Salim Roukos, Todd Ward, and Wei-Jing Zhu. 2002.
\newblock Bleu: a method for automatic evaluation of machine translation.
\newblock In \emph{Proceedings of the 40th annual meeting of the Association for Computational Linguistics}, pages 311--318.

\bibitem[{Radford et~al.(2019)Radford, Wu, Child, Luan, Amodei, Sutskever et~al.}]{radford2019language}
Alec Radford, Jeffrey Wu, Rewon Child, David Luan, Dario Amodei, Ilya Sutskever, et~al. 2019.
\newblock Language models are unsupervised multitask learners.
\newblock \emph{OpenAI blog}, 1(8):9.

\bibitem[{Rasouli et~al.(2020)Rasouli, Sun, and Rajagopal}]{rasouli2020fedgan}
Mohammad Rasouli, Tao Sun, and Ram Rajagopal. 2020.
\newblock Fedgan: Federated generative adversarial networks for distributed data.
\newblock \emph{arXiv preprint arXiv:2006.07228}.

\bibitem[{Reisizadeh et~al.(2020)Reisizadeh, Mokhtari, Hassani, Jadbabaie, and Pedarsani}]{reisizadeh2020fedpaq}
Amirhossein Reisizadeh, Aryan Mokhtari, Hamed Hassani, Ali Jadbabaie, and Ramtin Pedarsani. 2020.
\newblock Fedpaq: A communication-efficient federated learning method with periodic averaging and quantization.
\newblock In \emph{International conference on artificial intelligence and statistics}, pages 2021--2031. PMLR.

\bibitem[{Shahid et~al.(2021)Shahid, Pouriyeh, Parizi, Sheng, Srivastava, and Zhao}]{shahid2021communication}
Osama Shahid, Seyedamin Pouriyeh, Reza~M. Parizi, Quan~Z. Sheng, Gautam Srivastava, and Liang Zhao. 2021.
\newblock \href {https://arxiv.org/abs/2107.10996} {Communication efficiency in federated learning: Achievements and challenges}.
\newblock \emph{Preprint}, arXiv:2107.10996.

\bibitem[{Touvron et~al.(2023)Touvron, Lavril, Izacard, Martinet, Lachaux, Lacroix, Rozi{\`e}re, Goyal, Hambro, Azhar et~al.}]{touvron2023llama}
Hugo Touvron, Thibaut Lavril, Gautier Izacard, Xavier Martinet, Marie-Anne Lachaux, Timoth{\'e}e Lacroix, Baptiste Rozi{\`e}re, Naman Goyal, Eric Hambro, Faisal Azhar, et~al. 2023.
\newblock Llama: Open and efficient foundation language models.
\newblock \emph{arXiv preprint arXiv:2302.13971}.

\bibitem[{Vedantam et~al.(2015)Vedantam, Lawrence~Zitnick, and Parikh}]{vedantam2015cider}
Ramakrishna Vedantam, C~Lawrence~Zitnick, and Devi Parikh. 2015.
\newblock Cider: Consensus-based image description evaluation.
\newblock In \emph{Proceedings of the IEEE conference on computer vision and pattern recognition}, pages 4566--4575.

\bibitem[{Wang et~al.(2020)Wang, Yang, and Zhou}]{wang2020towards}
Cong Wang, Yuanyuan Yang, and Pengzhan Zhou. 2020.
\newblock Towards efficient scheduling of federated mobile devices under computational and statistical heterogeneity.
\newblock \emph{IEEE Transactions on Parallel and Distributed Systems}, 32(2):394--410.

\bibitem[{Wen et~al.(2023)Wen, Zhang, Lan, Cui, Cai, and Zhang}]{wen2023survey}
Jie Wen, Zhixia Zhang, Yang Lan, Zhihua Cui, Jianghui Cai, and Wensheng Zhang. 2023.
\newblock A survey on federated learning: challenges and applications.
\newblock \emph{International Journal of Machine Learning and Cybernetics}, 14(2):513--535.

\bibitem[{Wolf et~al.(2020)Wolf, Debut, Sanh, Chaumond, Delangue, Moi, Cistac, Rault, Louf, Funtowicz et~al.}]{wolf2020transformers}
Thomas Wolf, Lysandre Debut, Victor Sanh, Julien Chaumond, Clement Delangue, Anthony Moi, Pierric Cistac, Tim Rault, R{\'e}mi Louf, Morgan Funtowicz, et~al. 2020.
\newblock Transformers: State-of-the-art natural language processing.
\newblock In \emph{Proceedings of the 2020 conference on empirical methods in natural language processing: system demonstrations}, pages 38--45.

\bibitem[{Workshop et~al.(2022)Workshop, Scao, Fan, Akiki, Pavlick, Ili{\'c}, Hesslow, Castagn{\'e}, Luccioni, Yvon et~al.}]{le2023bloom}
BigScience Workshop, Teven~Le Scao, Angela Fan, Christopher Akiki, Ellie Pavlick, Suzana Ili{\'c}, Daniel Hesslow, Roman Castagn{\'e}, Alexandra~Sasha Luccioni, Fran{\c{c}}ois Yvon, et~al. 2022.
\newblock Bloom: A 176b-parameter open-access multilingual language model.
\newblock \emph{arXiv preprint arXiv:2211.05100}.

\bibitem[{Wu et~al.(2024)Wu, Liu, Wang, Li, Lv, Ling, Zhu, Zhang, Zheng, and Huang}]{wu2024advancing}
Muling Wu, Wenhao Liu, Xiaohua Wang, Tianlong Li, Changze Lv, Zixuan Ling, Jianhao Zhu, Cenyuan Zhang, Xiaoqing Zheng, and Xuanjing Huang. 2024.
\newblock Advancing parameter efficiency in fine-tuning via representation editing.
\newblock \emph{arXiv preprint arXiv:2402.15179}.

\bibitem[{Yang et~al.(2019)Yang, Liu, Chen, and Tong}]{yang2019federated}
Qiang Yang, Yang Liu, Tianjian Chen, and Yongxin Tong. 2019.
\newblock Federated machine learning: Concept and applications.
\newblock \emph{ACM Transactions on Intelligent Systems and Technology (TIST)}, 10(2):1--19.

\bibitem[{Zaken et~al.(2021)Zaken, Ravfogel, and Goldberg}]{zaken2021bitfit}
Elad~Ben Zaken, Shauli Ravfogel, and Yoav Goldberg. 2021.
\newblock Bitfit: Simple parameter-efficient fine-tuning for transformer-based masked language-models.
\newblock \emph{arXiv preprint arXiv:2106.10199}.

\bibitem[{Zeng et~al.(2022)Zeng, Liu, Du, Wang, Lai, Ding, Yang, Xu, Zheng, Xia et~al.}]{zeng2022glm}
Aohan Zeng, Xiao Liu, Zhengxiao Du, Zihan Wang, Hanyu Lai, Ming Ding, Zhuoyi Yang, Yifan Xu, Wendi Zheng, Xiao Xia, et~al. 2022.
\newblock Glm-130b: An open bilingual pre-trained model.
\newblock \emph{arXiv preprint arXiv:2210.02414}.

\bibitem[{Zhang et~al.(2022)Zhang, Roller, Goyal, Artetxe, Chen, Chen, Dewan, Diab, Li, Lin et~al.}]{zhang2022opt}
Susan Zhang, Stephen Roller, Naman Goyal, Mikel Artetxe, Moya Chen, Shuohui Chen, Christopher Dewan, Mona Diab, Xian Li, Xi~Victoria Lin, et~al. 2022.
\newblock Opt: Open pre-trained transformer language models.
\newblock \emph{arXiv preprint arXiv:2205.01068}.

\bibitem[{Zhang et~al.(2023)Zhang, Yang, Dai, Wang, Yu, Qu, and Xu}]{zhang2023fedpetuning}
Zhuo Zhang, Yuanhang Yang, Yong Dai, Qifan Wang, Yue Yu, Lizhen Qu, and Zenglin Xu. 2023.
\newblock Fedpetuning: When federated learning meets the parameter-efficient tuning methods of pre-trained language models.
\newblock In \emph{Annual Meeting of the Association of Computational Linguistics 2023}, pages 9963--9977. Association for Computational Linguistics (ACL).

\end{thebibliography}

\newpage

\appendix

\section{Selection of V}
\label{sec:appendix}

Regarding the specific choice of V, please refer to the appendix. The selection of V directly determines the quality of quantization. Based on QLoRA\citep{dettmers2024qlora}, it is observed that the model parameters follow a normal distribution. It is preferable for the standard numbers in V to be densely distributed in the middle range close to 0, while sparsely distributed towards the ends of the intervals at 1 and -1. This ensures a relatively even distribution of model parameters allocated to different standard numbers, leading to an optimal estimation of the original parameter matrix under the constraint of limited standard numbers. Specifically, when $w$=2, the selection of V in the FedLPP algorithm is $[-1.00, 0.00, 0.33, 1.00]$, and $V=[-1.00, -0.47, -0.21, 0.00, 0.16, 0.33, 0.56, 1.00]$ for $w=3$, In particular, when $w=1$, we choose $V$ as $[-1.00, 0.00, 1.00]$.

\section{Future Work}

Building upon the foundational success of \textsc{FedLPP}, several avenues for future research are evident to further refine and expand the capabilities of our FL framework. One immediate area of exploration involves the optimization of the quantization strategy to balance model performance with privacy preservation more effectively. Advanced techniques such as dynamic quantization or mixed-precision training could be employed to enhance model accuracy without compromising privacy.

Additionally, expanding the compatibility of \textsc{FedLPP} with various neural network architectures, including more complex models like GANs or transformers, could significantly broaden its applicability. Investigating the framework's effectiveness in these contexts will help in addressing the diverse needs of practical applications in different sectors such as healthcare, finance, and telecommunications.

Another promising direction is the exploration of hybrid approaches that combine \textsc{FedLPP} with other privacy-preserving techniques such as differential privacy or secure multi-party computation. Such combinations could offer layered security features, thereby providing stronger guarantees against potential privacy breaches.

Further, the development of resource management strategies to efficiently handle the computational and communication overheads in \textsc{FedLPP} would be crucial, especially for deployment in edge computing scenarios. Optimizing these aspects will ensure that the benefits of FL can be realized even in resource-constrained environments.

Lastly, conducting large-scale empirical studies to validate the framework's efficacy across different real-world datasets and environments would provide deeper insights into its practical implications and limitations. This would not only solidify the theoretical advancements made but also guide the practical implementations of FL systems.

\end{document}